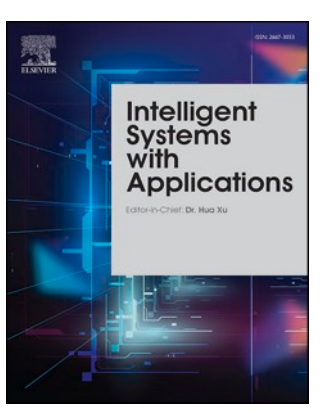

# Automated machine learning: AI-driven decision making in business analytics

Marc Schmitt

*Department of Computer Science, University of Oxford, UK*
*Department of Computer & Information Sciences, University of Strathclyde, UK*

ABSTRACT

The realization that AI-driven decision-making is indispensable in today's fast-paced and ultra-competitive marketplace has raised interest in industrial machine learning (ML) applications significantly. The current demand for analytics experts vastly exceeds the supply. One solution to this problem is to increase the user-friendliness of ML frameworks to make them more accessible for the non-expert. Automated machine learning (AutoML) is an attempt to solve the problem of expertise by providing fully automated off-the-shelf solutions for model choice and hyperparameter tuning. This paper analyzed the potential of AutoML for applications within business analytics, which could help to increase the adoption rate of ML across all industries. The H2O AutoML framework was benchmarked against a manually tuned stacked ML model on three real-world datasets. The manually tuned ML model could reach a performance advantage in all three case studies used in the experiment. Nevertheless, the H2O AutoML package proved to be quite potent. It is fast, easy to use, and delivers reliable results, which come close to a professionally tuned ML model. The H2O AutoML framework in its current capacity is a valuable tool to support fast prototyping with the potential to shorten development and deployment cycles. It can also bridge the existing gap between supply and demand for ML experts and is a big step towards automated decisions in business analytics. Finally, AutoML has the potential to foster human empowerment in a world that is rapidly becoming more automated and digital.

## 1. Introduction

AI-enabled intelligent systems are vital to survive and thrive in this era of relentless competition shaped by accelerating globalization and an ever-increasing basket of disruptive technologies (Warner & Wäger, 2019). All major industries – finance, healthcare, manufacturing, retail, supply chain, logistics, and utilities – have started to utilize advanced analytics and are potentially disrupted by applications based on machine learning (ML) and artificial intelligence (AI) (Dwivedi et al., 2021; Schmitt, 2020).

Data-driven decision-making based on AI/ML has become indispensable in today's global, fast-paced, and ultra-competitive market (Davenport, 2018). Business analytics plays a major part to facilitate this new way of decision-making (Schmitt, 2023). It is an interdisciplinary field and combines machine learning, statistics, information systems, operations research, and management science (Sharda, Delen & Turban, 2017) and is usually divided into descriptive, predictive, and prescriptive analytics (Delen & Ram, 2018).

The necessity to adopt sophisticated predictive models to make intelligent decisions is without question, but the ability to capture value through analytics is heavily dependent on employees with the required skill set to leverage those analytics capabilities (Grover, Chiang, Liang & Zhang, 2018). Even though initiatives toward data science education have started to manifest themselves (Clayton & Clopton, 2019), the huge demand for talent that makes sense of data and provides useful insights remains tremendous (Kar, Kar & Gupta, 2021). The use of non-experts when it comes to ML algorithms is problematic as extensive knowledge is required to successfully tune ML models (Schmitt, 2023).

Automated machine learning (AutoML) solutions have started to gain traction, which is a method to automatically tune and compare different algorithms to find the best hyperparameter combination (Feurer et al., 2015; He, Zhao & Chu, 2021). The preceding tasks of pre-processing and feature engineering of the dataset are only partly supported (Balaji & Allen, 2018), but the end goal of AutoML research is focused on automating the complete predictive modeling process. AutoML could help to fill the existing supply and demand gap when it comes to ML experts. It has also the potential to democratize ML across less quantitative academic disciplines and functional business areas to

*E-mail address:* marcschmitt@hotmail.de.

foster the creation of new research questions and business use cases. Several different AutoML solutions were introduced during the last few years. The major goal of the literature review was to choose the best suitable open-source AutoML framework for this study.

Gijsbers et al. (2019) offer an up-to-date comparison of the most mature open-source AutoML frameworks currently available: Auto-WEKA, auto-sklearn, TPOT, and H2O AutoML. The research itself is open-source and accessible online. It receives also regular updates upon the release of a new version. H2O AutoML is one of the top-performing models in this study.

Truong et al. (2019) analyzed the existing body of AutoML frameworks in terms of robustness and reliability taking into account a vast list of open-source and commercialized AutoML solutions. While there is no clear winner across all test cases, H2O managed to outperform all other models for regression and classification tasks.

This paper will zoom in on the predictive part of business analytics and analyze whether AutoML solutions can enhance the adoption rate of ML across business functions. Based on the literature review is the H2O AutoML framework the best choice for classification tasks and is hence the go-to framework for the following empirical study.

The objective of this study is to test whether the AutoML off the shelve frameworks have a similar performance and/or can beat manually trained ML models. This is important to further drive the adoption of ML solutions across business functions and domains as deep technical knowledge to develop new ML and DL models will require significant theoretical and technical training often not present in corporations. AutoML could speed up the development cycle and counteract the current skill shortage within the area and is the first step towards a full end-to-end decision engine for business analytics.

Whether this final step will materialize remains to be seen. Humans prefer to oversee decision-making; hence it is unlikely that all AI systems will run without any human input. Current literature in the area of intelligent systems has started to focus on human empowerment, including a stronger focus on augmentation instead of pure automation, which would be beneficial for humanity (Holmes et al., 2021; Maciej Serda et al., 2022; Toniolo et al., 2023). Overall, the goal is to expand the discussion in the hope to trigger new conversations and ultimately convincing more researchers to think about how to incorporate ML models within business processes. The ultimate goal is to empower humans and create a modern augmented workforce enabled by powerful data-driven decision toolkits based on intelligent systems.

In this study, the H2O AutoML framework is benchmarked against a manually created ML model to compare predictive ability and ease of use. Also, these findings will be used to discuss managerial implications for digital strategy. At last, a roadmap for future research will be presented.

The article is organized as follows. Section 2 “methods and materials” describes the AutoML framework used and the experimental design. Section 3 “numerical results” describes the outcome of the experiment and presents the performance of the H2O AutoML framework against the manually adjusted ML models. Section 4 discusses the numerical results, and managerial implications, and derives future research possibilities. Section 6 concludes with a summary.

## 2. Methods and materials

### 2.1. AutoML

Automated Machine Learning or AutoML is a method for automating the predictive analytics workflow. Depending on the concrete AutoML solution it might contain preprocessing, feature engineering, as well as model tuning. The current body of AutoML solutions does not handle the pre-processing very well (Truong et al., 2019) and the primary goal of this study is an assessment of the hyperparameter optimization and model choice. Based on the existing literature is the H2O AutoML framework one of the most mature AutoML solutions currently available. It achieves superior performance on classification and regression tasks according to several recent benchmark studies (Gijsbers et al., 2019; Truong et al., 2019). See Fig. 1.

The hyperparameter optimization of the different base models is done via a random grid search. It is a way of randomly selecting hyperparameter values from a fixed range of values. The system will then select combinations of these values uniformly at random. H2O offers the option of setting a stopping criterion, such as a maximum number of models or a maximum amount of time, to limit the search.

The stacked ensemble method of H2O is a supervised ensemble machine learning technique that uses a process called stacking to discover the optimal combination of a set of prediction algorithms (Algorithm 1). This process involves utilizing a meta-learner (specifically a non-negative GLM) to learn the best combination of the base learners (Ledell & Poirier, 2020). Concrete, H2O’s AutoML framework (H2O.ai, 2019) creates different candidate models such as GLMs, random forest, gradient boosting, and deep learning during an initial training phase and creates via stacking two different super learners. One super learner is based on all the pre-trained candidate models while the other is only an aggregation of the best model out of each family. The major parameters required for the AutoML solution are the *feature columns x*, the *response column y*, the *training_frame*, and the *validation_frame*. Also, the parameters max_*models* and max_*runtime_secs* are used to either specify the maximum number of models trained or the maximum time allowed for the process of model optimization. The H2O AutoML framework uses a random search as the optimization method.

### 2.2. Experimental design

The primary goal of this empirical study is to benchmark the H2O AutoML framework against a manually trained super-learning ensemble on real-world datasets from the domains credit risk, insurance claims, and marketing.

#### 2.2.1. Datasets

This experiment is based on three publicly available datasets to facilitate reproducibility and comparability. The chosen data sets can either be downloaded from the UCI machine learning repository or the public machine learning competition site Kaggle. The datasets contain 23, 57, and 16 features respectively, which are characteristics of the dataset (e.g., historical client data) and will serve as predictor variables to calculate the classification category of each observation. All three datasets contain a binary response column that identifies whether the client defaulted, initiates an insurance claim, or a marketing effort resulted in a sale. See Table 1 for a summary of the most important points of the case studies/datasets used in this study.

*2.2.1.1. Credit risk.* The first dataset stems from the credit risk domain and consists of payment information from credit card clients based in southeast Asia. The total dataset comes with 30,000 observations. 23,364 of them are categorized as positive cases and 6636 are negative cases – situations where the client defaulted. The observations contain 23 features including a response column that holds the binary default or non-default information. Concrete examples of the features present in this dataset are historical payment information, but also demographic information such as gender, age, marital status, and education.[1]

*2.2.1.2. Claims prediction.* The second dataset stems from the insurance domain and consists of information about automotive insurance policyholders. It has a total of 595,212 observations. A large portion of 573,518 of those observations are non-filed claims and 21,694 are filed

[1] The “Credit Risk” dataset can be accessed here: https://archive.ics.uci.edu/ml/datasets/default+of+credit+card+clients

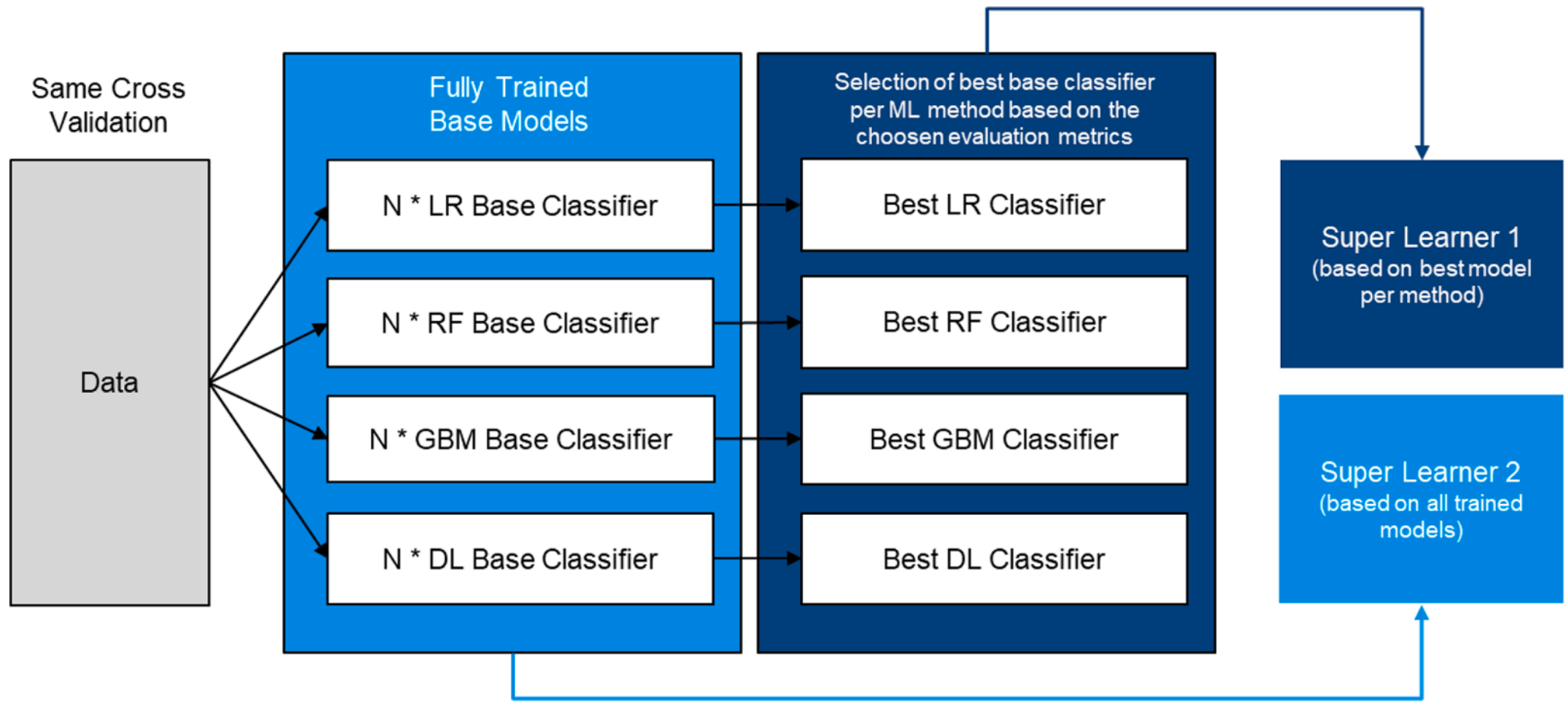

**Fig. 1.** The H2O AutoML framework trains several base learners and in a subsequent step combines those with two different super learners. One super learner is based on all previously trained classifiers, the other takes only into account the best classifier of each ML family (LR, RF, GBM, DL). H2O AutoML automatically ranks the outcomes based on the chosen evaluation metrics.

**Algorithm 1**
Pseudocode for Automated Machine Learning (AutoML).

**Input:** labeled test dataset Dt, labeled training dataset D1, number of cross-validation sets k, time to completion t, choice of meta-learner algorithm M
**Step 1:** Train Logistic Regression Classifier
**Step 2:** Train Deep Learning Classifier
**Step 3:** Train Gradient Boosting Machine Classifier
**Step 4:** Train Random Forest Classifier
**Step 5:** Use all pre-trained base classifiers to create super learner 1
**Step 6:** Use only the best classifier per category to create super learner 2
**Step 7:** Repeat steps 1–5 until the maximum number of models or time specified has been reached
**Output:** A list of classifiers build during the run-time in descending order based on their prediction accuracy on the test dataset Dt

**Table 1**
Description of datasets.

| Case Study | Observations | | | | | Description |
|---|---|---|---|---|---|---|
| | Total | $y = 0$ | $y = 1$ | Balanced* | Features | |
| Credit Risk | 30,000 | 23,364 | 6636 | 6636/6636 | 23 | Predict whether a customer will default on their loan payment |
| Claims Prediction | 5,95,212 | 5,73,518 | 21,694 | 21,694/ 21,694 | 57 | Predict whether a policyholder will file an auto insurance claim in the next year |
| Marketing | 45,211 | 39,922 | 5289 | 5289/5289 | 16 | Predict whether a target customer will open a checking account after a direct marketing effort |

* Random under-sampling was used to bring the datasets in a balanced state.

claims. The observations contain 57 features including a response column that indicates the claim status of each policyholder.[2]

*2.2.1.3. Marketing.* The third dataset holds marketing and sales data. It consists of customer information for a direct marketing campaign within the financial services domain. The total number of observations present in the dataset is 45,211. Out of those were 39,922 unsuccessful and 5289 were successful. Successful means resulting in a conversion or final sale. All observations come with 16 features and a response column indicating the binary outcome – has the marketing effort resulted in the final conversion/sale or not.[3]

### *2.2.2. Preprocessing*

Several preprocessing steps were necessary before running the experiment:

*2.2.2.1. Random undersampling.* A balanced distribution of positive and negative observations is necessary, otherwise, there will automatically be a natural pull to the class with the most observations, especially for data sets with a ratio of 90:10 or higher. This phenomenon would skew the results as the prediction accuracy may not come from the classifier but from the imbalance in the data set. This is achieved by random under-sampling. The goal here is to recalibrate the datasets and bring them into a state of equilibrium by eliminating observations of the majority class. Undersampling results in the loss of certain information, but that can be neglected here since the main purpose is to benchmark AutoML against the manually tuned super learner ensemble. An alternative would be to oversample the minority class, but this would blow up the dataset and increase training time – especially problematic for datasets that have already a significant amount of information.

*2.2.2.2. Encoding.* Another required preprocessing step for AI/ML models is the encoding of categorical data. Models in AI/ML require numeric input variables. Categorical data need to be transformed into a numerical version of themselves before model fitting and evaluation can take place. This is usually done by one-hot encoding (or ordinal encoding in case there exists already a rank). H2O contains a parameter setting called one_hot_explicit, which creates $N + 1$ new columns for categorical features with N levels.

*2.2.2.3. Training/Test split.* The chosen split is 80:20, which means 80% of the dataset will be used in the training process and the remaining 20% will be used to test the generalization ability of the trained classifier. The same cross-validation setup is necessary for the later fusion of the base learners into the meta learner. Hence, during the model training, 80% of the dataset will be split into different training and validation sets, which is done by cross-validation.

[2] The “Insurance Claims” dataset can be accessed here: https://www.kaggle.com/c/porto-seguro-safe-driver-prediction/data

[3] The „Marketing/Sales dataset can be accessed here: https://archive.ics.uci.edu/ml/datasets/Bank+Marketing

#### 2.2.3. Setup and evaluation

The inner workings of the AutoML solution offered by H2O train the 4 base-classifier Generalized Linear Model (LR), Random Forest, Gradient Boosting Machine, and Deep Feedforward Neural Networks. In a subsequent step, it applies the ensemble method stacking to fuse all those pre-trained candidate models to a super learner to increase the accuracy levels. The best model is automatically selected based on a chosen evaluation measure. To test the strength of this setup I have recreated the inner workings of the H2O AutoML solution by manually training the base models and combining them via stacking to a super learner. Overall, the comparisons are between two separately configured super learners. One is automatically generated by the H2O AutoML solutions, and one is manually tuned and configured. See Fig. 1 in 2.1. The four evaluation methods AUC, Accuracy, F-score, and LogLoss were used to assess the performance (Flach, 2019).

#### 2.2.4. Software

Preprocessing, model fitting, and evaluation are entirely conducted in RStudio, which is the integrated development environment (IDE) for the statistical programming language R (Core Team, 2019). R is one of the primary languages for data science and machine learning research and is also heavily used for prototyping in practice – especially for computational statistics. The R package H2O was used to set up the AutoML framework and also all the baseline models used for the experiment as Random Forest, Gradient Boosting Machines, and Deep Learning. H2O (LeDell & Gill, 2019) is an open-source machine learning platform written in Java and supports a wide range of predictive models. H2O has the advantage of speed as it allows us to move from a notebook/desktop-based environment to a large-scale environment. This increases performance and makes it easy to handle large data sets. H2O can be integrated into R-studio via a REST API.

## 3. Numerical results

In this section, the experimental results are presented. The manually tuned stacked ensemble learner was compared against the AutoML solution from H2O. Stacking was necessary to recreate the inner workings of the H2O AutoML procedure, which relies on training several different base classifiers including the subsequent combination of those pre-trained models for the final ensemble model based on stacking. Three real-world case studies in the areas of credit risk, insurance claims, and marketing were used in this experiment. The four evaluation matrices AUC, Accuracy, F-score, and LogLoss were used to benchmark the H2O AutoML solution against a manually optimized super learner. The Accuracy and F-score are reported at a 0.5 threshold level. The experiment was structured as follows:

In the first step the three baseline models random forest, gradient boosting machine, and deep learning were carefully trained. To tune the hyperparameter settings of the base models' traditional methods as grid search and random search over a pre-defined range of parameters, as well as manual adjustments, were used during the training process.

Table 2 shows the numerical results for the base classifiers for each dataset. Gradient Boosting obtained the highest overall performance, followed by Random Forest. Deep Learning has the lowest performance scores. This is consistent across all three datasets.

In the second step, the candidate models were combined with a so-called super learner via the ensemble method stacking that has been proven to deliver asymptotically optimal improvements upon a set of base classifiers. For each case study, all three base models (RF, GBM, DL) were used to create the super learner. All three combinations of the baseline models for the stacked ensemble were tested and the best performance could be achieved by using RF, GBM, and DL as input for the super learner for all three case studies. This is not always the case.

**Table 2**
Numerical results of optimized base classifiers for all three case studies.

| Case Study | Method | AUC | Accuracy | F-score | Logloss |
|---|---|---|---|---|---|
| Credit Risk | Random Forest | 0.769 | 0.708 | 0.683 | 0.574 |
| | Gradient Boosting | 0.775 | 0.716 | 0.694 | 0.570 |
| | Deep Learning | 0.758 | 0.703 | 0.686 | 0.609 |
| Claims Prediction | Random Forest | 0.636 | 0.598 | 0.584 | 0.667 |
| | Gradient Boosting | 0.640 | 0.598 | 0.586 | 0.663 |
| | Deep Learning | 0.633 | 0.597 | 0.534 | 0.669 |
| Marketing | Random Forest | 0.940 | 0.877 | 0.885 | 0.318 |
| | Gradient Boosting | 0.940 | 0.878 | 0.886 | 0.299 |
| | Deep Learning | 0.933 | 0.864 | 0.871 | 0.322 |

In the last step, the stacked super learner created in step two serves as a benchmark for the AutoML solution from H2O to evaluate its performance, robustness, and reliability. Table 3 shows the final comparison of the H2O AutoML solution and the trained super learner.

Overall, the results are surprisingly consistent and the stacked Super Learner was able to outperform the AutoML model on all three datasets with an AUC difference of 0.002.

While performance deltas for the other matrices are not identical, the stacked ensemble outperformed the AutoML solution here as well in most cases. For the credit risk case study, the difference is 0.003 for Accuracy, 0.003 for F-score, and 0.002 for LogLoss. The performance difference in the case of the insurance dataset is 0.004 for Accuracy, 0.002 for F-score, and 0.001 for LogLoss. The performance difference for the marketing case study is −0.001 for Accuracy, −0.002 for F-score, and 0.001 for LogLoss. AutoML slightly outperformed the stacked ensemble only in the marketing case study in terms of Accuracy and F-score. Overall, the manually tuned stacked ensemble shows superior performance compared to the AutoML solution for all three case studies.

## 4. Discussion

The purpose of the experimental study presented in this paper was to test the performance of the H2O AutoML framework compared to a manually tuned ML model in terms of the four evaluation measures AUC, Accuracy, F-score, and LogLoss. This section has three parts: First, the results of the empirical study will be discussed to assess the overall performance of the tested AutoML solution. Second, the findings will be discussed w.r.t. to business analytics to better understand the implications of those findings for managers, practitioners, and researchers. And last, a roadmap for future research is provided.

### 4.1. Discussion of results

In a nutshell – based on the findings of the empirical study on three real-world datasets from the domains of credit risk, insurance, and marketing – the H2O AutoML model was not able to outperform the manually tuned classifier. It had difficulties reaching the quality of a manual setup in two ways:

**Table 3**
Comparisons of the super learner benchmark model and AutoML for all three case studies.

| Case Study | Method | AUC | Accuracy | F-score | Logloss |
|---|---|---|---|---|---|
| Credit Risk | Stacked Ensemble | 0.778 | 0.717 | 0.698 | 0.565 |
| | AutoML | 0.776 | 0.714 | 0.695 | 0.567 |
| Claims Prediction | Stacked Ensemble | 0.642 | 0.603 | 0.592 | 0.662 |
| | AutoML | 0.640 | 0.599 | 0.590 | 0.663 |
| Marketing | Stacked Ensemble | 0.944 | 0.883 | 0.889 | 0.299 |
| | AutoML | 0.942 | 0.884 | 0.891 | 0.300 |

(1) The underlying models (based classifiers) did not reach the same prediction accuracy as the manually tuned versions. Increasing the running time did not have a significant impact on the final output and did not result in a performance improvement.
(2) The H2O AutoML package chooses two stacked ensemble combinations. One is based on all the trained models and the other is based on the best model for each category. It does not test whether another combination of the candidate models (e.g., a smaller subset) results in better performance. This is important as adding weaker models to the total pool of models for the stacked ensemble unnecessarily sabotages the performance. Only the best baseline models should be considered for the super learner as additional classifiers tend to dilute the performance by adding non-optimal information that results in a reduction of prediction accuracy. This was also demonstrated by Guo, He and Huang (2019) and (Schmitt, 2020).

However, the performance delta is not very strong and the AutoML solution provided by H2O is a potent model tuning engine that can significantly speed up prototyping or help practitioners less familiar with ML concepts to set up a powerful model. Nevertheless, for maximum prediction accuracy, careful model tuning and adjustments of hyperparameters done by a data scientist result in the best performance. Based on the small performance improvement it is questionable whether the small edge of manual adjustment as demonstrated by the three case studies can account for the time-consuming model creation process when almost the same can be achieved with no knowledge and adjustment efforts. The answer to this question is mainly dependent on the use case at hand, and whether a tiny performance improvement justifies the additional time required for manual model tuning. Also, given the strong performance of the AutoML solution created by H2O, it is almost certain that further research will result in prediction accuracy levels that are on par with models adjusted by ML experts.

Overall, AutoML is an important first step toward complete end-to-end decision processes. Due to its relatively strong performance, and consistent results, AutoML has the potential to become more capable and hence a valuable tool for human engineers over time. This would significantly help to democratize ML for business analytics functions, especially for small to medium-sized businesses, which tend to have more difficulties to hire the appropriate talent.

Practical Note: Model performance can change drastically depending on the data that is used for training (Schmitt, 2020). The concrete accuracy (e.g., 70%, 80%, or 90%) is largely related to the underlying dataset. For example, the literature in the field of credit scoring makes it clear that every credit risk dataset has different accuracies (Schmitt, 2022). Better model tuning or different model choices have the potential to improve prediction accuracy, but only in a small range. In practice, we need to work with the dataset we have (or improve it) and choose the best model that is available. If the trained model improves upon the current working solution in terms of prediction accuracy, it is an improvement over the benchmark. AutoML can help as a starting point but cannot – in its current state – replace a team of human experts in those situations. In addition, the business context needs to be considered. For example, credit decisions depend only partially on the probability of default (PD). It is possible to reduce the loss given default (LGD) with insurance or collateral. This reduces the impact of the PD accuracy since the amount lost during the default for the lender would be reduced or zero. While AI/ML has become vital to remain competitive, it is important to remember that it is often only a part of a more complex system.

### 4.2. Managerial implications

Management has always used data to generate information for insights. Mainly in the form of business information systems. This is not new. However, the earlier more intuitive business approach gradually changed towards evidence-based and data-driven decision-making (Brynjolfsson & Mcelheran, 2019; Delen & Ram, 2018). This new form of decision-making requires an environment capable of utilizing the power of artificial intelligence and machine learning.

AutoML is a big first step and might gradually evolve and extend to a fully automated decision engine. It has the potential to create a new level playing field by democratizing ML solutions across industries and business lines. Even though the findings in this study prove that AutoML does not yet beat careful human engineering when it comes to model tuning, it could help to support the adoption of ML solutions by helping to fill the talent gap. In addition, it is useful to support skilled data scientists with fast prototyping and benchmarking, which could lead to accelerated development cycles and earlier deployment. The findings in this study are a strong indication that AI/ML solutions will get less cost-intensive and more user-friendly over time due to continued innovation within the field itself as well as due to hardware improvements, better software, APIs, and UIs.

As these developments continue domain knowledge and subject matter expertise will likely be more important to develop and implement end-to-end AI solutions compared to expertise in machine learning itself. Agrawal, Gans and Goldfarb (2019) argue that domain expertise cannot be commoditized, but ML as a general-purpose technology for decision-making can and will be commoditized in different ways. How long this process will take is difficult to predict, but the commoditization of AI/ML solutions has already started and can be observed in the real world. Predominantly large cloud providers such as AWS, Google Cloud, and MS Azure are continuously optimizing their AI/ML platforms, which can conveniently be purchased via the software as a service (SaaS) model.

Fully automated ML solutions pose the potential to democratize analytics across several industries and business functions which could lead to tremendous value gains. However, one problem with the adoption of big data and advanced analytics models in business is decision-makers with a tendency to focus on bottom-line results and a need for visible business value (Kushwaha, Kar & Dwivedi, 2021). Clear communication with a focus on value realization is needed to establish a data-driven culture in corporations.

Nevertheless, AutoML is not yet able to automatically preprocess complex datasets, which is one of the most time-consuming steps in the data science process. The same is true for the need to move from pure predictive outputs to concrete actionable steps in the form of prescriptive analytics. Until the last steps towards a complete end-to-end process are not solved corporations need to rely on hiring data science experts or external consultants to help them drive the current digital transformation initiatives.

In addition, developments in the field of AI and automation have caused many employees to worry that they will be replaced by AI technology, and subsequently avoid using it. We should also not forget that employees desire to be given more autonomy in their work and to have the freedom to make their own decisions (Maciej Serda et al., 2022). Hence, a move towards an augmented workforce, where modern analytics tools support human decision making is more favorable. We should imagine and strive for a future where powerful intelligent systems team up with humans for co-creation, which would foster human empowerment in a world that is rapidly becoming more automated and digital.

### 4.3. Future research

Further research is required at both ends of the predictive analytics process and also when it comes to human empowerment and augmentation. AutoML needs to be able to handle data preprocessing to further automate the ML pipeline. Also, in the end, when it comes to deriving concrete actions from those predictions there is room for improvement. However, the notion of complete off-the-shelf ML solutions for data-driven decision-making might not be the ideal end goal. It might be more desirable to strive for an augmented workforce, where intelligent systems are highly integrated with the human workforce.

#### 4.3.1. Preprocessing

The first step of supervised predictive analytics, which can take up to 80 or 90 percent of the required time, is widely considered to be the most important aspect of data science. This step involves different kinds of data cleaning and adjustments but also entails transforming the feature space into one that is more informatively conclusive. It has been observed that superior feature engineering can win online data science competitions such as Kaggle, suggesting that this initial phase is more critical than the model-tuning process itself. Research exists on automated feature engineering; however, machines cannot always semantically categorize related features as well as domain experts. This is a valid future research direction.

#### 4.3.2. Prescriptive analytics

Current research mainly focuses on predictive tasks and results must be interpreted by human decision-makers. One of the most interesting questions in business analytics is how to move from predictive analytics to a complete end-to-end decision engine that provides managerial decision-makers with concrete actions that can be acted upon. So far, ML and DL are predominately used for predictive analytics. There are already attempts to combine ML methods with operations research/ management science to move from pure predictions to actual decisions, but how to go from a good prediction to a good decision is poorly understood. The major problem is to account for uncertainty in the decision-making process (Bertsimas & Kallus, 2019). Looking at recent studies in other domains such as Alphastar show that this is possible and that deep reinforcement learning (DRL) can reach human-level decision power in an uncertain environment and in real time (Vinyals et al., 2019). However, studies on DRL for prescriptive analytics and managerial decision-making in an uncertain environment do not yet exist and would open several new research questions within the fields of business analytics, intelligent systems, and information management. This is a domino that needs to fall to reach full end-to-end decision processes within business analytics.

#### 4.3.3. Human in the loop

Augmentation instead of full automation is becoming increasingly popular in business analytics (Davenport, 2018). It may be more beneficial to integrate human expert knowledge at this early stage instead of completely automating it. This could allow for better control and feedback loops and help to understand ML models and their predictions better. AI and humans should be able to interact with each other effortlessly. A human-in-the-loop setup that takes advantage of both AI and human capabilities could indeed be the ideal solution (Toniolo et al., 2023). The integration of AI with human ingenuity is a promising area of future research.

## 5. Conclusion

The continuous digitalization of our world economy resulted in an increased demand for experts in the field of machine learning and artificial intelligence. This increased demand has led to a skill shortage, which slowed down the adoption of AI/ML methods in business analytics. AutoML frameworks have the potential to narrow this current talent gap and could also accelerate the predictive analytics process. The H2O AutoML framework in its current capacity does not reach the full prediction accuracy that is possible by careful manual adjustment of the models. However, despite those findings, this study has shown that AutoML can be a powerful tool. First, it can be used as a baseline during prototyping for ML experts, which can help to accelerate the development and deployment cycles of ML projects; second, it makes ML models more accessible to non-expert users as it further increases user-friendliness by moving the level of abstraction higher; third, AutoML can be considered as a big step towards the construction of a full end-to-end decision engine in business analytics; and finally, AutoML could help to foster human empowerment by creating an augmented workforce. We should aim for a future where intelligent systems and people team up to work together, which will promote human control over a world that is increasingly digitalized and automated.

## Author contributions

This research article is the sole work of the author listed.

## Declaration of Competing Interest

The authors declare that they have no known competing financial interests or personal relationships that could have appeared to influence the work reported in this paper.

## Data availability

Data will be made available on request.